\documentclass[11pt]{article}

\usepackage[final]{acl}
\usepackage{tabularx}
\usepackage{array}
\usepackage[table]{xcolor}
\usepackage{booktabs}
\usepackage{makecell}
\usepackage{amsmath}
\usepackage{colortbl}
\usepackage[dvipsnames]{xcolor}
\usepackage{url}
\usepackage{multicol}  
\usepackage{tcolorbox} 
\usepackage{booktabs}
\usepackage{makecell} 
\usepackage{multirow} 
\usepackage{alltt}
\usepackage{amssymb}
\usepackage{siunitx}
\usepackage{paralist}
\usepackage{pifont}
\usepackage{courier}
\newcommand{\cmark}{\ding{51}} 
\newcommand{\xmark}{\ding{55}} 
\newcommand{\oneS}{\ensuremath{{}^{\textstyle *}}}

\usepackage[ruled,vlined]{algorithm2e}
\RestyleAlgo{boxruled} 
\newcommand{\showfontsize}{%
  \fbox{size=\the\dimexpr\f@size pt\relax,\ baselineskip=\the\baselineskip}%
  \typeout{*** Font size: \f@size pt ; baselineskip: \the\baselineskip}%
}
\makeatother

\title{Interactive In-Meeting Speaker Correction with Human Feedback}

\author{
 \textbf{Xinlu He\textsuperscript{1}},
 \textbf{Yiwen Guan\textsuperscript{1}},
 \textbf{Badrivishal Paurana\textsuperscript{1}},
\\
 \textbf{Pitipat Kongsomjit\textsuperscript{1}},
 \textbf{Zilin Dai\textsuperscript{1}},
 \textbf{Jacob Whitehill\textsuperscript{1}}
\\
\\
 \textsuperscript{1}Worcester Polytechnic Institute
\\
 \small{
   \textbf{Correspondence:} 
   \href{mailto:jrwhitehill@wpi.edu}{jrwhitehill@wpi.edu}
 }
}

\begin{document}
\maketitle
\begin{abstract}
Most automatic speech processing systems operate in ``open loop'' mode  without user feedback about who said what, yet human-in-the-loop workflows can potentially enable higher accuracy. We propose an LLM-assisted in-meeting speaker correction system that lets users fix speaker attribution errors through brief corrective feedback. After performing streaming ASR and diarization, the system presents concise LLM-generated summaries to help users identify important speaker errors, and it incorporates user feedback by updating the speaker-attributed transcript and adding  online speaker enrollments.
To make this workflow effective despite errors in speech processing, LLM analysis, and user feedback, we developed several mechanisms to identify the intended correction more precisely. Further, we built  an LLM-driven user feedback simulation to evaluate the workflow reprodubilty and at scale. Applied to the AMI headset test set, our system substantially reduces the DER from a streaming baseline (Google ASR + ECAPA) by 31.99\% and speaker substitution error by 52.68\%. Results of a pilot usability study suggest several avenues  to improve the user experience.
\end{abstract}

\section{Introduction}
\label{sec:intro}
Although state-of-the-art automatic speech recognition (ASR) systems have made substantial progress~\cite{bain2023whisperx, qwen}, accurate transcription  remains challenging in many realistic conditions, including noisy backgrounds, long-tail vocabulary, and overlapping speech. Beyond word recognition errors, speaker attribution errors can also undermine transcript reliability by obscuring who said what.~\cite{ryant2021diharddiarizationchallenge,he2025surveyendtoendmultispeakerautomatic}. Subsequent correction is thus often necessary to obtain high accuracy~\cite{radhakrishnan2023whisperingllamacrossmodalgenerative,llm_speech_correction}. However, in standard workflows, any such correction is usually performed post hoc, incurring costly manual annotation. A more practical approach in some settings is to enable the speakers to make lightweight, natural corrections to speaker-attributed text during the conversation. This is akin to how people naturally correct misunderstandings by their interlocutors during a meeting. Recent advances in large language models for interactive workflows~\cite{llm_interaction1_2024,peng2025surveyspeechlargelanguage} open new possibilities for this approach to succeed.

In this paper, we propose an in-meeting interactive correction system (Fig.~\ref{fig:system}) that enables users to provide online feedback to correct the speaker attribution errors. The pipeline performs streaming automatic speech recognition and diarization to produce a speaker-attributed transcript, then uses an LLM to summarize the content for the users. During the conversation, users can provide brief corrective feedback. The system immediately incorporates this feedback, identifies the intended speaker-attributed text span, updates the transcript, and uses the corrected speech as online enrollment evidence for future speaker assignment.

In designing this workflow, we adhered to three constraints that are important for practical use: (1) The user should not be required to read an overwhelming amount of information (e.g., an entire transcript), as this would be both distracting and impractical; (2) All feedback should be lightweight (e.g., just a simple verbal correction, not highly detailed timing information), so as to minimize intrusiveness; and (3) the ASR and diarization must be streaming (i.e., they must make a decision immediately, without waiting for more context at the end of the session).  Creating an effective workflow is thus quite challenging. 

To make feedback-based correction effective despite errors in the ASR, diarization, LLM outputs, we devised several mechanisms to help the workflow to perform corrections at a finer granularity than the original ASR segments. Since ASR segments may contain speech from multiple speakers, we first partition speech segments into more speaker-consistent units using an efficient segmentation step. The LLM then narrows each user correction to the intended speaker-attributed text span, so that transcript updates and enrollment evidence are based on the specific error being corrected rather than an entire ASR segment.
We evaluate the workflow with an LLM-driven user feedback simulation on the AMI headset test set. Finally, we present a pilot usability study to explore how real-time correction impacts the dialog flow and to uncover which design features could improve the user experience the most.

{\bf Contributions}:
(1) We propose an LLM-assisted in-meeting workflow for interactive speaker correction, where users provide simple feedback messages to correct speaker attribution errors during a conversation.
(2) We design a fine-grained speaker correction mechanism that maps user feedback to the intended speaker-attributed text span and uses corrected spans as online enrollment evidence for future speaker assignment.
(3) We develop an LLM-driven user feedback simulation for scalable evaluation on the AMI test set. Experiments show that our system reduces DER by 31.99\% and speaker error (SErr) by 52.68\%.
(4) We conduct a pilot study to understand the factors that impact the user's experience when using the proposed workflow.

\section{Related Work}
\label{sec: related work}

\textbf{Feedback-based Correction }

Prior work has studied user- and model-assisted correction of ASR transcripts.
Early work~\cite{kolkhorst-etal-2012-evaluation} showed that simple user correction leads to significant readability improvements. More recent approaches leverage user feedback at scale: \cite{2023_ASRU_Feedback_ASR} takes the feedback in the ASR model to correct transcription using federated learning.
\cite{johnson25_interspeech} introduced an LLM-assisted human-in-the-loop pipeline where a small set of manual fixes drives large-scale error correction. Relatedly, LLM-assisted subjective annotation highlights how human guidance can reliably steer model outputs with reduced labeling cost~\cite{schroeder-etal-2025-just}. 

These works mainly address lexical transcription errors and are often offline or post-hoc.
Although recent work has begun to explore interactive ASR correction~\cite{wang2026interactiveasrhumanlikeinteraction}, it focuses on correcting recognized content rather than speaker attribution.
Our work instead studies real-time speaker-attribution correction in meeting assistants.

\textbf{LLM-enabled Human-AI Interaction}

Recent advances in large language models (LLMs) have shifted the paradigm of human-AI interaction toward fluid, feedback-driven mediation between user intent and system behavior~\cite{Zou2025LLMBasedHC, Gao2024Taxonomy}. 

Our system uses LLMs in two ways. First, LLM-generated summaries provide lightweight context for user correction: prior work shows that feedback to conversational agents is often limited by common-ground, verifiability, and communication barriers~\cite{10.1145/3772318.3791600,axelsson2022modeling}, while LLM meeting recap systems use highlights and structured minutes to support efficient review~\cite{summariesLLMmeeting}.
We therefore use LLM-generated glanceable summaries to help users notice errors and provide voice feedback.
Second, feedback must become executable system updates: prior work on structured output studies how to constrain LLM responses for downstream workflows~\cite{10.1145/3613905.3650756, xie2023translating}. We similarly translate natural-language feedback into structured speaker-reassignment actions for transcript updates and online enrollment.

\section{Proposed Workflow}
\label{sec:pagestyle}
\begin{figure*}[t]
\begin{center}
    \includegraphics[width=1\textwidth]
    {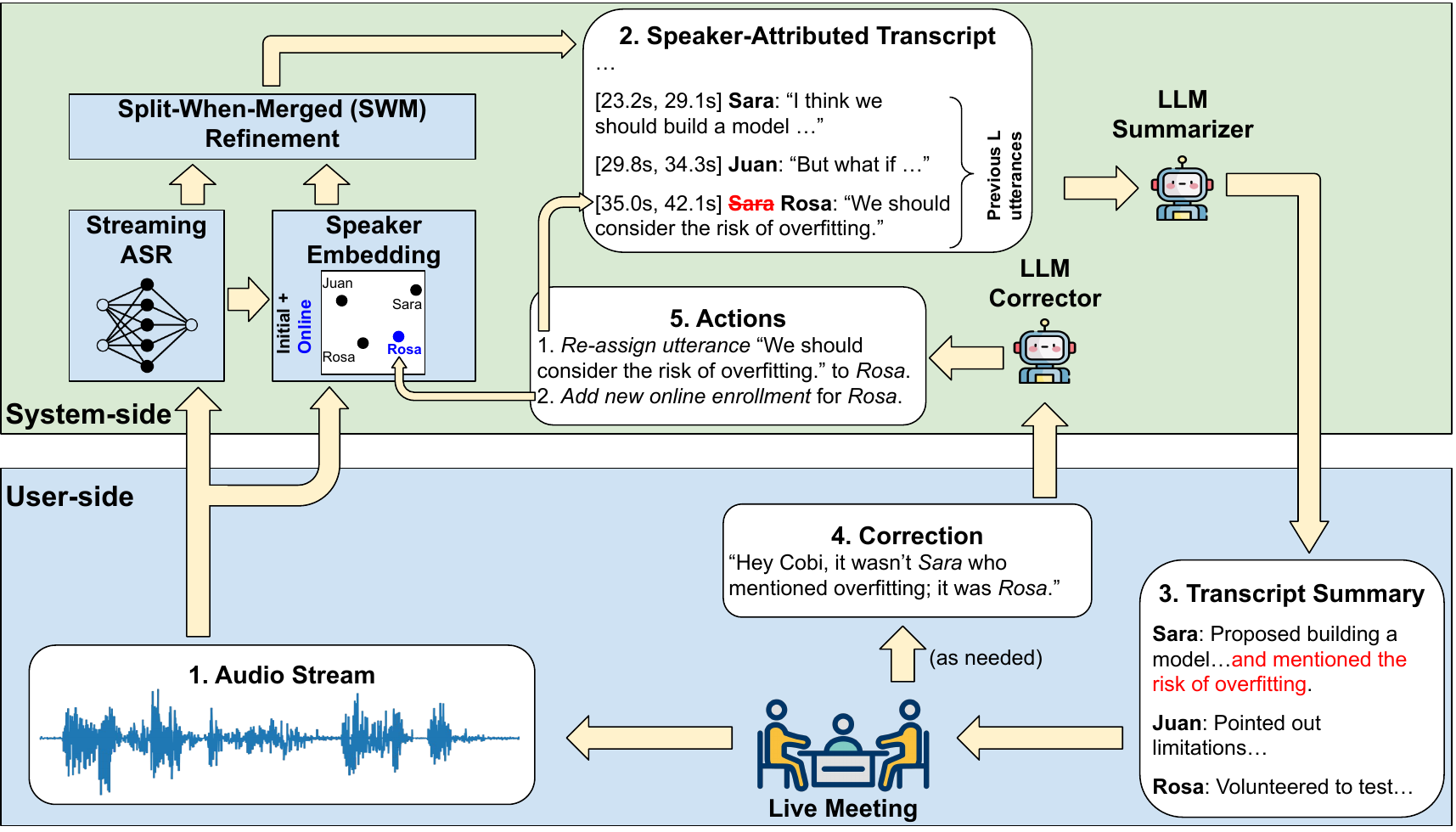}
    \end{center}
    \caption{\textbf{Proposed LLM-enabled workflow}. Streaming audio is transcribed and assigned speaker labels; the most recent $L$ turns are summarized for users; and verbal feedback is parsed by an LLM corrector to update the targeted speaker-attributed span and online enrollment. 
    }
    \label{fig:system}
\end{figure*}

\label{subsec:system overview}

Fig.~\ref{fig:system} illustrates the overall interactive speaker correction workflow for real-time multi-speaker meetings. We assume that each speaker has enrolled themself at the start of the meeting by providing a short sample of their speech (a few seconds long) along with their name. On the system side, the workflow is modular and harnesses off-the-shelf tools such as an LLM (e.g., GPT-4.1), streaming automatic speech recognition (we used Google Speech-to-Text API, but alternatives such as Parakeet RNN-T are possible), and speaker embedding model (e.g., ECAPA-TDNN). 


{\bf Transcription \& Segmentation}: First, the live audio stream (box \#1 in the figure) is processed by a streaming ASR to produce a sequence of segments, \(\mathcal{S}=\{S_1,S_2,\dots\}\), where each segment $S_t$ has an overall start and end time, a list of words, as well as the start and end times of each word. Although each segment often corresponds to a complete sentence spoken by a single speaker, in practice it might be just a sentence fragment or a mix of words from multiple speakers. Note that, since we assume a streaming setting, neither the ASR nor the diarizer has access to information from the future conversation. 

{\bf Attribution}: From each segment $S_t$, we use a speaker embedding model to extract a speaker embedding from the corresponding speech signal based on the segment's start and end times. This embedding is compared using cosine similarity with the set of all enrollments -- both the initial enrollments  as well as a set of online enrollments (described below) -- to infer the speaker of the segment.
This results in a speaker-attributed transcript (\#2). Optionally, the speaker-attributed segment can be  refined using the Split-When-Merged technique (see Section \ref{sec:tass}).

As shown in the figure, the speaker-attributed segmentation can contain mistakes, such as mis-attributing the sentence ``We should consider the risk of overfitting'' to Sara (rather than the true speaker, Rosa).

{\bf Summarization}: 
In real-world meetings, the transcript grows very quickly, is difficult to parse, and would be impractical for a human speaker to monitor without becoming completely distracted from the actual conversation. (Indeed, during pilot experiments, we were quickly overwhelmed by the running transcript of the ASR when we tried to both speak naturally and monitor the ASR's output.)
Hence, in order to provide speakers with a useful but efficient representation, we use an LLM Summarizer (see the exact prompt in the Appendix) to condense it into a summary (\#3) of the most recent $L$ segments. Specifically, the summary is designed to communicate just the main contributions of each speaker to the recent conversation. This running summary can be shown continually to the users and provides a quick glance as to what the speech system believed to have been communicated.

{\bf Correction}: 
During the live meeting, a user may occasionally look at the summary and realize that something about the transcription was incorrect. They might then wish to correct the transcript by providing feedback, either manually (e.g., by typing in a chat interface) or verbally (e.g., using a wake-word to initiate a speech correction).
While such feedback could be both about the content (what was said) and the attribution (who said it), we focus on just the latter. For example, if a speaker notices that Sara was erroneously credited with mentioning the risk of overfitting, they might provide a correction (\#4) such as ``Hey Cobi, it wasn't Sara who mentioned overfitting; it was Rosa,'' where  ``Cobi'' is our system's wake-word. In the case of spoken feedback, the correction sentence can be detected by the ASR and then separated from the ``actual'' conversation by detecting the wake-word.

{\bf Remediation}: 
The speaker's feedback is processed by the LLM Corrector (see prompt in the Appendix) to determine what remedial actions (\#5) should be taken. 
In particular, our system performs two kinds actions: \emph{Re-assign} a segment in the transcript to the true speaker (Sara $\rightarrow$ Rosa, in our example), and \emph{add new online enrollment} for the true speaker. In particular, the online enrollment uses the start and end times of the identified segment to extract a new speaker embedding; if the duration of the segment is within pre-specified bounds (we chose 5 to 20sec), then the online enrollment for the correct speaker is added to the embedding space (see the blue dot for Rosa in the figure). This can thus help to prevent future attribution mistakes, both by giving the speaker embedding model more reliable information on each speaker, and also by helping it to overcome domain shift (e.g., if the background noise changed significantly over the conversation). Note that simply asking the speakers to provide multiple enrollment utterances up-front can be more burdensome and is often unnecessary for speakers whom the embedding model already recognizes accurately. Instead, our workflow adds additional enrollments dynamically based on mistakes that are flagged by the user.


\subsection{Affordances and Challenges}
With the proposed workflow, speech attribution errors in the recent past can be fixed by users themselves through simple, unobtrusive feedback. Moreover, due to the online enrollments, which give the system a better sense of what each speaker sounds like, future errors can be prevented as well. Adaptive feedback based on actual attribution errors are also more flexible and efficient than, say, simply asking all users to provide multiple enrollments up-front.

On the other hand,  it is not trivial to design the workflow to be both practical in terms of user experience and effective for improving transcript accuracy. There are many things that can go wrong, and thus it was not obvious a priori whether it would improve diarization accuracy. Below we describe some of the challenges involved:

{\bf No tedious manual feedback}: Manually providing the exact moment when an error occurred, and specifying how to correct it (e.g,. using the mouse to click-and-drag on a timeline from the speaker-attributed transcript to fix the segmentation) would be prohibitively tedious for the users and completely disrupt the natural conversation flow. Instead, we rely on simple feedback messages that can be provided either verbally (through speech) or through a chat interface. This is much more convenient but also less precise. 

{\bf Speech segmentation errors}: Streaming ASR front-ends typically segment audio based on voice activity cues rather than accurate speaker turns. Hence, a single sentence of actual speech might sometimes span multiple segments, or a single segment might comprise words from multiple speakers. This makes the job of the LLM Corrector more difficult when it has to determine remedial actions based on the user's feedback.
 
{\bf Bad summarization}: We found that even strong LLMs (e.g., GPT-4o) can make subtle mistakes when summarizing a running transcript, even when the latter is itself correct. In turn, an incorrect summary can result in erroneous feedback from the speakers.
    
{\bf Wrong feedback}: 
Speakers' corrections could be wrong, ambiguous, or incorrectly parsed by the system itself (particularly when
issued verbally rather than manually). 

{\bf Incorrect remediation}: Assuming the speaker's feedback is correct (based on the summary received), we found that even a strong LLM may infer false remedies to
``fix''  mistakes, e.g., re-assigning the wrong segment from the transcript to a particular speaker.
    
{\bf Misleading online enrollments}: If the speech segment used to provide the online enrollment is not accurate (e.g., wrong start/end times), or if it is not ``representative'' of the speaker's voice, then it may actually degrade, rather than enhance, the speaker attribution.

Tackling these challenges required careful workflow design, LLM prompt engineering (which we performed on our training set, so as to avoid overfitting), and extra processing to refine the speech segmentation and user's correction handling.


\begin{table*}[t]
\centering
\setlength{\tabcolsep}{4pt}
\renewcommand{\arraystretch}{1.08}
\fontsize{8.0}{9}\selectfont

\begin{tabular}{p{0.25\linewidth} p{0.75\linewidth}}
\toprule
\multicolumn{2}{c}{\textbf{Example of Split-When-Merged (SWM) and Narrowed LLM Correction}} \\
\midrule

\rowcolor{gray!10}
\textbf{Reference} &
\textbf{FEO072}: I can explain it. \\
\rowcolor{gray!10}
&
\textbf{FEO070}: Yeah. \\
\rowcolor{gray!10}
&
\textbf{MEE071}: Have you annotated your code nicely? \\
\rowcolor{gray!10}
&
\textbf{FEO072}: Oh yeah. Actually, in the moment it's got a load of rubbish because the the \\
\addlinespace[3pt]

\rowcolor{red!8}
\textbf{Google ASR + ECAPA} &
\textbf{FEO070}: I can explain it yeah have you annotated your code nice oh yeah actually in the moment it's got a load of rubbish because the the \\
\addlinespace[3pt]

\rowcolor{blue!8}
\textbf{After SWM} &
\textbf{MEE071}: I can explain it yeah have you annotated your code nice \\
\rowcolor{blue!8}
&
\textbf{FEO070}: oh yeah actually in the moment it's got a load of rubbish because the the \\
\addlinespace[3pt]

\textbf{Correction} &
Hey COBI, \textit{MEE071} didn't say they could explain it, \textit{FEO072} did. \\

\addlinespace[3pt]

\rowcolor{green!8}
\textbf{Corrected Transcript} &
\textbf{FEO072}: I can explain it \\
\rowcolor{green!8}
&
\textbf{MEE071}: yeah have you annotated your code nice \\
\rowcolor{green!8}
&
\textbf{FEO070}: oh yeah actually in the moment it's got a load of rubbish because the the \\

\bottomrule
\end{tabular}

\caption{
Example from AMI meeting EN2002a. The baseline ASR segment merges multiple speakers; SWM splits it into more speaker-consistent units; narrowed LLM correction further fixes the error by changing only the target span from \textit{MEE071} to \textit{FEO072}.
}
\label{tab:qualitative_swm_correction}
\end{table*}

\subsection{Fine-Grained Speaker Correction}
\label{subsec:fine-grained speaker correction}
As discussed above, a streaming ASR can sometimes produce a speech segment that incorrectly combines speech from multiple speakers; this can happen  when speakers talk in quick succession in turn transitions. This mismatch makes whole-segment speaker attribution unreliable. An example from the AMI speech dataset, when processed by the Google Cloud Speech-to-Text API, is shown in Table \ref{tab:qualitative_swm_correction}; the ASR incorrectly combines four segments into just a single segment and then incorrectly assigns it to speaker FEO070.

We mitigate this issue
in two steps: First, we add a post-processor, Split-When-Merged (SWM), to the ASR segments
that uses acoustic information to refine each segment into smaller, speaker-consistent sub-segments using word-level speaker evidence. Second, given user feedback, the LLM uses semantic information to locate the specific part of a segment that are relevant, rather than correcting the entire segment. See Table \ref{tab:qualitative_swm_correction} for an example.


\subsubsection{Split-When-Merged Refinement}
\label{sec:tass}

We devised Split-When-Merged (SWM) as an efficient, training-free, post-ASR method to split  speech segments at token boundaries so as to increase the ``purity'' of speaker assignment within each subsegment. We found that it significantly increased the accuracy of the speaker-attributed transcript and  made the user feedback easier for the workflow to parse.

Given an ASR segment $S$ with word-level timestamps, SWM recursively performs binary partitioning. 
Since speaker embeddings extracted from very short audio clips (e.g., a single word) can be unreliable, we adopt a sliding window strategy and determine each word's speaker label through voting across overlapping windows.
Specifically, using sliding windows of length $W$ and stride $\Delta$, we obtain speaker votes over time by extracting a speaker embedding from each window and comparing it to each speaker enrollment (both the initial and online enrollments). Each word $k$ is assigned to the speaker $y_k$  that receives the majority of votes across all windows overlapping its temporal span.

Next, we evaluate the segment's speaker ``purity'' $\rho(S)$ as the maximum count of words assigned to a single speaker 
$\rho(S)=\max_s \sum_{k=1}^{n}\mathbf{1}[y_k=s]$,
where $s$ represents a speaker, $n$ is the number of words in the segment, and $\mathbf{1}$ is the indicator function.
If $\rho(S)/n \ge \theta$, where $\theta$ is a purity threshold, we keep the segment unchanged.
Otherwise, we evaluate all candidate binary split points $i \in \{1,\ldots,n\}$ to maximize the speaker consistency on both sides:
$\Phi(i)=\rho(S[1:i]) + \rho(S[i+1:n])$.
The optimal boundary is determined by $i^\ast=\arg\max_i \Phi(i)$.
The split is accepted only if the dominant speakers of the two resulting sub-segments differ. 
The partition  recurses until all sub-segments satisfy the purity threshold $\theta$.

\subsubsection{Narrowing the Correction}
In our workflow, the user feedback is expected to correct the attribution of a meaningful sentence or phrase, while ASR segments may span multiple sentences, possibly from multiple speakers.
Therefore, applying corrections or adding online enrollments at the whole-segment level can over-correct unrelated words and introduce noise into the  online enrollments  We thus found it useful, in terms of the workflow's accuracy, to include instructions in the prompt of the  LLM Corrector to sub-select only the relevant parts of the corrected segment. In particular, given the SWM-refined speaker-attributed transcript, user correction feedback, and the speaker list as input, the LLM Corrector produces a JSON-like structure output, including the target sentence in the segment, the original speaker, and the corrected speaker. The LLM acts as a semantic matcher between natural-language feedback and the transcript. 

\section{User Feedback Simulation}
\label{sec:simulation}
To evaluate the proposed workflow in a way that is reproducible and that does not require large-scale human-in-the-loop annotation, we simulated a user study by generating speaker feedback using an LLM.  This allows us to estimate the benefit of the workflow on standard speech datasets while keeping the feedback format close to natural user instructions. For simplicity, the feedback generated by our simulator is textual (e.g., given through a chat interface) rather than verbal, but this could be extended using a text-to-speech engine.

In our simulation pipeline, we assume the speakers know the ground-truth speaker-attributed transcript of who said what during the conversation. Based on this knowledge as well as the transcript summary they receive from the interactive workflow
(box \#3 in Figure \ref{fig:system}), they may sometimes give a correction (as illustrated in \#4).

In particular, after every $L$ segments, the system generates a summary, and we simulate potential user feedback.  Each simulation consists of three LLM calls: (1) An LLM compares the transcript summary with the corresponding ground-truth transcript to determine whether there is a speaker-attribution error and identify the error if one exists. (2) If an error is identified, the LLM generates a natural-language correction request that mimics user feedback starting with the wake words ``Hey Cobi.''. (3) 
An additional check  is applied to verify that the proposed correction is reasonable before passing it to the system. 
For each meeting, we limit the total number of simulated corrections across all speakers to $C_{\max}$.

{\bf Simulation Fidelity}:
Even with a strong LLM such as GPT-4o, simulating accurate corrections based on the ground-truth transcript and the transcript summary is non-trivial.
To assess the accuracy of the simulated user feedback itself,
we manually examined and annotated the simulated corrections on 8 meetings from the AMI training set. 
With the refinements described in Appendix~\ref{sec:appendix Prompt Tuning}, the simulated feedback reaches 78.9\% correction accuracy on the training dataset. While it is not a perfect substitute for real user feedback, it provides a reasonably accurate and scalable proxy for evaluating the downstream correction pipeline on public meeting datasets.

\label{subsec:simulation}

\section{Experiments}
\label{sec:typestyle}
\begin{table*}[t]
\centering
\small
\begin{tabular}{lrrrrrr}
\toprule
\multirow{2}{*}{\textbf{System}} 
& \multicolumn{2}{c}{\textbf{Actual}} 
& \multicolumn{2}{c}{\textbf{Improvement}} 
& \multicolumn{2}{c}{\textbf{Oracle Bound}} \\
\cmidrule(lr){2-3}\cmidrule(lr){4-5}\cmidrule(lr){6-7}
& DER$\downarrow$ 
& SErr$\downarrow$
& ImDER$\uparrow$
& ImSErr$\uparrow$
& DER$\downarrow$
& SErr$\downarrow$ \\
\midrule
Baseline (Google ASR + ECAPA)
& 36.32 & 22.05 
& -- & -- 
& $^{\star}$23.74 & $^{\star}$9.48 \\

Baseline (Google ASR + ECAPA) + SWM
& 31.60 & 17.33 
& 13.00 & 21.40 
& $^{\star}$19.51 & $^{\star}$5.25 \\
\midrule
Up-front Enrollments 
&27.23 & 12.96
&25.03 & 41.21
& $^{\dagger}$20.55 & $^{\dagger}$6.28 \\

Up-front Enrollments (Matched Count)
& 28.92 & 14.65
& 20.39 & 33.58
& $^{\dagger}$20.55 & $^{\dagger}$6.28 \\
\midrule
Proposed workflow (Correction Only)
& 31.94 & 17.67
& 12.06 & 19.86
& $^{\dagger}$20.55 & $^{\dagger}$6.28 \\
Proposed workflow (Online Enrollment Only)
& 26.82 & 12.55
& 26.16 & 43.08
& $^{\dagger}$20.55 & $^{\dagger}$6.28 \\
\textbf{Proposed workflow} 
& \textbf{24.70} & \textbf{10.43} 
& \textbf{31.99} & \textbf{52.68} 
& $^{\dagger}$20.55 & $^{\dagger}$6.28 \\


\bottomrule
\end{tabular}
\caption{Main results on the AMI test set. The first group reports uncorrected baselines; the second uses ground-truth upfront enrollments (``Matched Count'' matching real enrollment count in proposed system); and the last group shows ablations of the proposed workflow. $^{\star}$ and $^{\dagger}$ denote the all-segment correction oracle and the one-additional-enrollment-per-speaker oracle, respectively. All values are percentages.}
\label{tab: main_results}
\end{table*}
Using user feedback simulations, we conducted experiments  to assess whether the proposed workflow can improve diarization accuracy. Since our workflow 
is for in-meeting
interaction, we assume that the ASR and diarizer are \emph{streaming} and thus cannot access information in the future; our results are thus not comparable with offline diarization pipelines.

We compare our workflow to multiple alternative strategies: (1) Baseline:  streaming ASR and diarization system such as Google ASR + ECAPA-TDNN (i.e., the speaker-attributed transcript shown in box \#2 in Figure \ref{fig:system}), either with or without the SWM refinement step; (2) a strong, non-interactive workflow in which each speaker provides multiple ``up-front'' enrollments of themselves at the start of the meetings. Importantly, the latter approach assumes precisely identified start and end times of each enrollment, which would need to be tediously input manually. Finally, to estimate the upper bound on how much the diarization accuracy could possibly improve with any workflow that fixes speaker attribution, we compare to (3) an oracle that uses the same segment boundaries as the baseline but optimally assigns them to the correct speaker.

\subsection{Dataset}
We evaluated on the Headset-mix test set of the AMI Meeting Corpus \cite{carletta2005ami} (16 meetings, about 9 hours; 14-50 minutes each, 21.34 minutes on average; 3-4 speakers per meeting), which is a standard benchmark for multi-speaker ASR and diarization. 

\subsection{Evaluation Metrics}
We compute Diarization Error Rate (DER) with pyannote 
\cite{pyannoteMetrics}
including missed speech, false alarm, and speaker confusion.
In our workflow, ASR detection is fixed, and correction only affects speaker attribution; therefore, missed speech and false alarms remain unchanged. We report the speaker error (SErr) separately along with relative improvements in DER and SErr over the baseline, denoted ImDER and ImSErr. In all experiments, we use a 0.25s collar and set \texttt{skip\_overlap=True}, because each ASR-derived segment is assigned to one speaker.


\subsection{Implementation Details}
We used GPT-4o for summarization, correction, and simulated user-feedback generation. Exact prompts are in Appendix~\ref{fig:appendix_prompt}. ASR was performed with Google  Speech-to-Text in streaming mode to obtain segment- and word-level timestamps; speaker embeddings were extracted with ECAPA-TDNN~\cite{ECAPA}. SWM used 1.0s windows with 0.2s stride, at most one recursive split per segment, and purity threshold $\theta=0.7$. Summaries are shown every $L=10$ utterances. 
Simulated corrections are limited to $C_{\max}=30$ per meeting.






\section{Results and Analysis}

Table~\ref{tab: main_results} summarizes the main results. 

\textbf{SWM}
Compared with the uncorrected baseline,  split-when-merge (SWM)  reduces DER from 36.32\% to 31.6\%, and SErr from 22.05\% to 17.33\%, showing that finer segmentation improves speaker attribution. SWM also improves the oracle bound by reducing oracle SErr from 9.48\% to 5.25\%.

\textbf{Proposed Workflow DER}
The proposed workflow further improves speaker attribution accuracy on top of the SWM technique.  With one online enrollment per speaker, it achieves the best performance: 24.70\% DER and 10.43\% SErr, corresponding to 31.99\% and 52.68\% relative improvements over the uncorrected baseline. Since our correction  changes speaker labels rather than segment boundaries, the only reducible component in DER is SErr. Relative to the gap between the uncorrected SErr of 22.05\% and the SWM oracle bound of 5.25\%, our method closes about 70\% of the gap.

\textbf{Online vs Upfront Enrollment}
We further compare online enrollment with up-front enrollment. 
Although the up-front enrollment condition use ground-truth speaker segments from the beginning of the meeting, they are less effective than the proposed online enrollment strategy. With one additional up-front enrollment per speaker, DER is 27.23\% with 12.96\% SErr.
With the same number of online enrollments as the interactive system (at most 1), up-front enrollment achieves 28.92\% DER and 14.65\% SErr, while the proposed workflow reduces them to 24.70\% and 10.43\%. 
This suggests that online enrollment is more useful than relying only on initial speech samples, likely because they are collected from actual attribution errors and reflect the acoustic conditions encountered later in the meeting.

\textbf{Benefit of Online Enrollments}
We examined the benefit of the online enrollments above and beyond the benefit of the corrective feedback itself. The online enrollments reduced the DER more substantially than just the corrections themselves (compare the last 3 lines of Table \ref{tab: main_results}). In turn, this reduces the need by the users to give feedback: In our simulations, the Correction Only condition triggers one simulated correction every 88s on average, while the full proposed workflow (including online enrollments) triggers one every 104.5s. 
In the full workflow, each speaker issues 3.1 corrections in one meeting on average.

\textbf{Comparison to SoTA  Diarization}
Recent streaming diarization systems  report DERs on the AMI dataset in the range of 20\% to 30\%~\cite{2025benchmarkingdiarizationmodels,gruttadauria2024onlinespeakerdiarizationmeetings,lightweight_AVSR,10.1007/978-3-031-40498-6_16,kwon2022absolutedecisioncorruptsabsolutely,coria2021overlapawarelowlatencyonlinespeaker}. These works, which are often not open source, are based on technical improvements to the underlying diarization backbone.
In contrast, our focus is not to propose a new  backbone, but to study an interactive  workflow based on human feedback. The ASR and diarizer in our system are therefore replaceable, and stronger streaming diarization or attribution modules could be integrated in future work.



\subsection{Ablation Studies}

\textbf{SWM and Online Enrollment}
Table~\ref{tab:split_ae_intervals} reports relative improvements over the baseline for different combinations of SWM and online enrollment (OE). We test $H_0:\,\mathbb{E}[\mathrm{ImDER}]\leq 0$ with one-sample $t$-tests over all testing 16 meetings. Without either component, the system does not improve over the baseline. Using either component alone yields significant gains: SWM-only achieves 12.06\% ImDER and 19.86\% ImSErr, while OE-only achieves 15.28\% ImDER and 25.20\% ImSErr. Combining them gives the largest gains, reaching 31.99\% ImDER and 52.68\% ImSErr.  All three non-empty variants are significant ($p<0.05$), suggesting that SWM and OE are complementary: SWM provides cleaner speaker-consistent units for OE to improve future speaker assignment.

\begin{table}[t]
\centering
\setlength{\tabcolsep}{5pt}
\begin{tabular}{ccccl}
\toprule  
\textbf{SWM} & \textbf{OE} 
& ImDER$\uparrow$ & ImSErr$\uparrow$ & $p$-value \\
\midrule
\xmark & \xmark  & -0.25\%  & -0.41\%  & 0.6862 \\
\cmark & \xmark & 12.06 \% &19.86 \%  & 0.00002\oneS\\
\xmark  & \cmark  & 15.28\%  & 25.20\% &  0.00496\oneS\\
\cmark & \cmark & \textbf{31.99\%}  & \textbf{52.68\%} & 0.00004\oneS\\
\bottomrule
\end{tabular}
\caption{Ablation on SWM and OE. \oneS indicates $p<0.05$ (one-sample $t$-test across meetings).}
\label{tab:split_ae_intervals}
\end{table}
\begin{table}[t]
\centering
\resizebox{\linewidth}{!}{
\begin{tabular}{ccccccc}
\toprule
\multirow{2}{*}{\textbf{$L$}} &
\multicolumn{3}{c}{\textbf{Conversation}} &
\multicolumn{3}{c}{\textbf{Summary}} \\
\cmidrule(lr){2-4}\cmidrule(lr){5-7}
& \makecell{Avg.\\ words} & \makecell{Im.\\ DER$\uparrow$} & \makecell{Im.\\ SErr$\uparrow$}
& \makecell{Avg.\\ words} & \makecell{Im.\\ DER$\uparrow$} & \makecell{Im.\\ SErr$\uparrow$} \\
\midrule
\textbf{5} &80&25.96&42.75&29&31.53&51.94\\
\textbf{10} &153&\textbf{29.60}&\textbf{48.76}&38&\textbf{31.99}&\textbf{52.68}\\
\textbf{15} &216&26.46&43.58&45&23.79&39.17\\
\textbf{20} &281&20.59&33.9&50&26.54&43.72\\
\bottomrule
\end{tabular}}
\caption{Conversation vs.\ Summary across context length $L$.}
\label{tab:conv_vs_summary_by_interval}
\end{table}

\textbf{Summarization vs Conversation Correction}
Table~\ref{tab:conv_vs_summary_by_interval} compares full conversation display with LLM-generated summaries. All results are with SWM and one OE. Summary-based correction achieves stronger gains than conversation-based correction for $L\in \{5,10,20\}$, while using  fewer words per window.
The summary view remains compact, increasing from 29 to 50 words, whereas the conversation view grows from 80 to 281 words. This suggests  concise summaries provide enough context to identify  attribution errors while reducing the reading load. 

\section{Pilot User Study}
%
To complement the quantitative analyses, we also
conducted a pilot usability study among 3 members of our laboratory (co-authors of this paper) to examine the practical usability of the pipeline. As
these participants are obviously not impartial, we used this study to identify ways to improve the user experience and
to generate specific hypotheses that could be studied in future work. In particular,
we examined (1) how much time a user requires to read a summary and (possibly) give a correction; (2) whether users ``acclimate'' to interacting with the system; and (3) general factors that impact the users' experience.
To this end, we created  a software prototype using an accelerated version of Whisper  and ECAPA. It uses the LLM prompts for summarization and correction as described in Sec.~\ref{subsec:system overview} and supports real-time correction (invoked by the system's wake-word, ``Hey Spark\ldots'') but does not include the fine-grained speaker correction mechanisms from Sec.~\ref{subsec:fine-grained speaker correction}. 

Our study consisted of 3 sessions (10m39s, 17m34s, and 22m46s, resp.), during which the participants freely discussed different topics: (1) interactive speech recognition, (2) artificial intelligence's impact on society, and (3) education. During each session, the participants occasionally checked the summary and sometimes issued a verbal correction (e.g., ``Hey Spark, the comment about making new jobs was from Jacob''). The goal was to interact with the system naturally so that it would accurately document who contributed which ideas. All participants enrolled themselves (for 5s) prior to  the first session.  After conducting the sessions, we then analyzed the saved audio file, transcripts, and system logs.

{\bf Findings}: Over all 3 sessions, there were 13 corrections, which together took 3m58s (about 7.8\% of the total study time). Of these, the prototype correctly detected 9 of them based on the wake-word. This suggests even a weak ASR (like the \begin{tt}whisper-tiny\end{tt} model we used) can detect corrections most of the time. The median time spent by the users in giving a correction was 18.5s (min 7s, max 35s). While it is difficult to judge whether this is a ``large'' amount of time, we note that in many online research meetings the participants spend significant time manually taking notes (e.g., via a shared Google document) and discussing them.
We also
found that both the maximum and median time spent giving corrections decreased from session 1 (max 35s, med 19s) to session 3 (max 20s, med 18s), suggesting that users might become more efficient at giving corrections.

We observed opportunities for improvement: (1) Real-time responsiveness on our  laptop (AMD Ryzen 7 7840u CPU) requires small models, which  are less accurate. In turn, this decreases the quality of the speaker-attributed transcript and hampers the ability of the LLM Corrector to process user feedback.  (2) Non-English names are often not be recognized correctly, which inhibits accurate correction when naming speakers.  (3) The LLM Summarizer might benefit from a customized prompt depending on the meeting goals, e.g., summarize the key ideas vs.~record only the action items. (4) The choice of the wake-word can matter. We initially tried ``Cobi'' but found that ``Spark'' was detected more reliably.

\section{Conclusion}
We presented a LLM-assisted framework for in-meeting interactive correction of speaker-attributed transcripts using lightweight user feedback.
The framework combines fine-grained correction with online enrollment to correct  speaker attribution errors in transcripts.
Using an LLM-driven user simulation, the system reduces DER on AMI by 31.99\% and speaker error by 52.68\%, showing that precise, adaptive feedback can substantially improve speaker-attributed transcription.
\section*{Limitations}
Our evaluation uses LLM-simulated user feedback in the experiment rather than live user studies. While this enables scalable evaluation on AMI, it may not fully capture the real situations.
Also, our current system focuses on speaker-attribution correction. As future work, the same interactive framework could be extended beyond speaker attribution to support lexical ASR correction and more direct handling of boundary or overlap errors.

\section*{Ethical Standards}
This work is a machine learning study on previously collected data, and thus no ethical approval was required.

\bibliography{refs}
\appendix
\section{Prompt Tuning}
\label{sec:appendix Prompt Tuning}

\begin{table*}[t!]
\centering
\setlength{\tabcolsep}{4pt}
\renewcommand{\arraystretch}{1.1}
\small
\begin{tabular}{p{0.25\linewidth} p{0.75\linewidth}}
\toprule
\multicolumn{2}{c}{\textbf{ ASR Error Example in ES2002d (S9)}} \\
\midrule

\rowcolor{blue!6}
\textbf{Ground Truth} &
\textcolor{ForestGreen}{\textbf{MEE008}}: \textbf{How about with embossing the logo isn't that going to cost us some money} \\
\rowcolor{blue!6}
&
FEE005: Doesn't say so \\
\rowcolor{blue!6}
&
MEE008: Yeah, that's a freebie \\
\rowcolor{blue!6}
&
\textcolor{red}{\textbf{MEE007}}: Reckon that probably counts as a special form for the buttons \\
\rowcolor{blue!6}
&
MEE006: Yeah \\
\rowcolor{blue!6}
&
FEE005: Yeah that's a good idea \\
\addlinespace[3pt]

\rowcolor{blue!3}
\textbf{ASR Output} &
\textcolor{red}{\textbf{MEE007}}:\textbf{ with embossing the logo isn't that going to cost us some money} does it say so is that for your account is our special forum yeah yeah that's a good idea \\
\addlinespace[3pt]

\rowcolor{orange!8}
\textbf{Simulated Feedback} &
Hey COBI, \textcolor{red}{\textbf{MEE007}} didn't raise concerns about \textbf{the cost of embossing the logo}, it was actually \textcolor{ForestGreen}{\textbf{MEE008}}. \\
\addlinespace[3pt]

\rowcolor{red!5}
\textbf{Corrected Output V1} &
\textcolor{ForestGreen}{\textbf{MEE008}}: with \textbf{embossing the logo} isn't that going to cost us some money does it say so is that for your account is our special forum yeah yeah that's a good idea \\
\addlinespace[3pt]

\rowcolor{green!7}
\textbf{Corrected Output V2} &
\textcolor{ForestGreen}{\textbf{MEE008}}:\textbf{ with embossing the logo isn't that going to cost us some money}\\
\rowcolor{green!7}
&
\textcolor{red}{\textbf{MEE007}}: does it say so is that for your account is our special forum yeah yeah that's a good idea \\

\midrule
\multicolumn{2}{c}{\textbf{ Summarization Error Example in ES2002c (S19)}} \\
\midrule

\rowcolor{blue!6}

\rowcolor{blue!3}
\textbf{ASR Output} &
\textcolor{ForestGreen}{\textbf{FEE055}}: about any the clever check of the case is flat the single curve in this \textbf{double curved}\\
\rowcolor{blue!3}
&
MEE054: um\\
\rowcolor{blue!3}
&
\textcolor{red}{\textbf{MEE054}}: I'm not exactly sure what these \textbf{give it like a slightly more aesthetic feel and a double curve} when required to perform Miracles with the plastic when\\
\addlinespace[3pt]

\rowcolor{purple!6}
\textbf{Summary} &
\textcolor{ForestGreen}{\textbf{FEE055}}: ... and mentioned \textbf{showing something related to a double curve}.  

... 

\textcolor{red}{\textbf{MEE054}}: ... and \textbf{questioned the aesthetic impact of a double curve}.\\
\addlinespace[3pt]
\rowcolor{orange!8}
\textbf{Simulated Feedback} &
Hey COBI, \textcolor{red}{\textbf{MEE054}} didn't question the aesthetic impact of a \textbf{double curve}, it was \textcolor{ForestGreen}{\textbf{FEE055}}. \\
\addlinespace[3pt]

\midrule
\multicolumn{2}{c}{\textbf{ Simulation Error Example in ES2002d (S31)}} \\
\midrule

\rowcolor{blue!6}

\rowcolor{blue!3}
\textbf{ASR Output} &
\textcolor{red}{\textbf{MEE056}}: \textbf{remote control} I've never seen 1 of them before they're all batteries I think so\\
\rowcolor{blue!3}
&
MEE054: voice recognition especially not 1 where you could\\
\rowcolor{blue!3}
&
\textcolor{ForestGreen}{\textbf{MEE053}}: yeah you got voice recognition computers it's not \textbf{remote controls} yeah\\
\addlinespace[3pt]

\rowcolor{purple!6}
\textbf{Summary} &
\textcolor{ForestGreen}{\textbf{MEE053}}: Discussed a project involving debugging computer software and addressing machine crashes.  


...

\textcolor{red}{\textbf{MEE056}}: Commented on the high-tech nature of products on a website and mentioned \textbf{remote controls with batteries}.\\
\addlinespace[3pt]
\rowcolor{orange!8}
\textbf{Simulated Feedback} &
Hey COBI, \textcolor{red}{\textbf{MEE056}} mentioned remote controls with batteries, but it was actually \textcolor{ForestGreen}{\textbf{MEE053}}.\\
\addlinespace[3pt]

\bottomrule
\end{tabular}

\caption{An example of ASR concatenation error. Multiple turns are merged into one ASR segment, leading to incorrect speaker attribution.ES2002d S9}
\label{tab:error_example}
\end{table*}
LLMs are used in both the designed system and the simulation experiments. In the system, the LLM acts as a summarizer and a corrector. In simulation, the LLM is initially used for two functions: one to analyze the differences between the ground-truth conversation and the generated summary by the system, and the second function is to simulate the correction utterance beginning with 'Hey Cobi'. 

We tune the prompts by example learning and error analysis. All prompt tuning and analysis are conducted only on the training set (8 meetings).
The common errors observed in the initial analysis include:

\begin{itemize}
\item \textbf{ASR/VAD Error:} Even with SWM, speaker-turn boundaries remain difficult for current streaming ASR tools with VAD. As a result, a short utterance from speaker A may be concatenated with a longer utterance from speaker B into a single segment. Although speaker A's contribution is short, it may still contain the key content and appears in A's summary. In this case, correcting the whole segment without acoustic evidence can further degrade diarization result. 

To address this issue, we modify the system's corrector's prompt so that the LLM is instructed to correct only the portion of the utterance that is explicitly referenced in the user-provided correction. The downstream system then revises the speaker assignment only for that span, while leaving the remaining portion unchanged according to the original acoustica attribution. The first example in Table~\ref{tab:error_example} shows such an error caused by ASR/VAD segmentation. After the prompt refinement, the system  tends to isolate the semantically corrected span instead of reassigning the entire merged utterance.

\item \textbf{Summarization Error:} In conversational interactions, multiple speakers may discuss the same topic, and the LLM summarizer may generate overly generic summaries that fail to capture the subtle distinctions between speakers. Moreover, interruptions, statement continuation, and short reactions between speakers are common in natual interactions. The LLM  summarizer may supply and completes the whole sentence with the context from other speakers, which further increases difficulty for raising accurate corrections later.

To address this issue, we refine the LLM summarization prompt to encourage more specific words, and discourage highly similar summaries across speakers.For important opinions, the summarizer is instructed to attribute the content only to the primary speaker expressing the idea, rather than to speakers who merely agree or provide short follow-up reactions. Example 2 in Table~\ref{tab:error_example} shows an instance of this error type in preliminary experiments.

\item \textbf{Simulation Error:} Simulating user correction feedback is particularly difficult when multiple speakers mention the same keyword or discuss the same topic. In such cases, the generated correction may appear plausible at the topical level while still being incorrect in speaker attribution. Even with prompt tuning and separating difference analysis process and the generation process, this issue remains. Importantly, this challenge arises in the simulation experiments, which is used to simulate user feedback, rather than in the interactive system itself. 
\end{itemize}

To improve the simulation in a simple and practical way, we introduce an additional verification step after feedback generation to check whether a proposed correction is speaker-attributionally reasonable. This verifier is implemented in a zero-shot setting and is used only for simulation quality control. In addition, following \cite{leviathan2025promptrepetitionimprovesnonreasoning}, we repeat the same prompt twice, which yields more stable and reliable output.


In preliminary experiments, we observed that the LLM is generally effective as the system corrector for locating the corresponding utterance in the predicted transcript and revising its speaker attribution. The main source of errors, instead, lies in generating incorrect speaker-correction proposals. 

Table~\ref{tab:correction_breakdown} summarizes the total number of corrections, the proportion of correct versus incorrect corrections, and the distribution of error types before and after the prompt and pipeline refinements. After refinement, the total number of correction proposals decreases from 82 to 57, mainly due to the addition of proposal verification. At the same time, the proportion of correct corrections increases substantially, from 36.6\% to 78.9\%. Across error types, summarization errors are fully eliminated. ASR/VAD-related errors are reduced by roughly half, from 20.7\% to 10.5\%, owing to the partial-span correction strategy. Simulation errors are reduced by about two-thirds, from 29.3\% to 10.5\%, primarily due to the added verification step.

\begin{table}[t]
\centering
\small
\begin{tabular}{lcc}
\toprule
\textbf{Category} & \textbf{Original} & \textbf{Final} \\
\midrule
\rowcolor{gray!12}
Total Corrections & 82 & 57 \\
\rowcolor{green!12}
\quad Correct & 30 (36.6\%) & 45 (78.9\%) \\
\rowcolor{red!12}
\quad Wrong & 52 (63.4\%) & 12 (21.1\%) \\
\rowcolor{red!6}
\quad\quad ASR Error & 17 (20.7\%) & 6 (10.5\%) \\
\rowcolor{red!6}
\quad\quad Summary Error & 11 (13.4\%) & 0 (0.0\%) \\
\rowcolor{red!6}
\quad\quad Simulation Error & 24 (29.3\%) & 6 (10.5\%) \\
\bottomrule
\end{tabular}
\caption{Breakdown of LLM-generated correction suggestions on the training set. Total suggestions are divided into correct and wrong cases, and wrong cases are further categorized by error type. Percentages are computed with respect to the total number of suggestions in each setting.}
\label{tab:correction_breakdown}
\end{table}

\section{Prompts}
\label{sec:appendix_prompt}
\begin{figure*}[t]

\begin{tcolorbox}[
  colback=white,
  colframe=black,
  boxrule=0.3pt,
  arc=0pt,
  width=\textwidth,
  left=2.5mm,
  right=2.5mm,
  top=1.5mm,
  bottom=1.5mm,
  boxsep=0.8mm
]
\fontsize{8}{9}\selectfont
\begin{alltt}
### TASK DESCRIPTION
You are summarizing a meeting transcript for speaker attribution analysis. 

### INPUT
Transcript: "conversation"

### INSTRUCTIONS
Instructions:
- Write ONE short sentence for each speaker. Use the speaker IDs exactly as they appear. 
- The summary can be not complete. All the words and phrases in one speaker's summary
should be from the same speaker in the transcript, not from context of other speakers.
- Contain important keywords in summary. Avoid repeating or rephrasing all sentences
from the transcript. 
- Ignore short backchannel responses that do not contain meaningful information 
(e.g., "yeah", "okay", "uh-huh", "I agree")

### OUTPUT REQUIREMENT
Give a short summary by speaker. Make it simple to read and understand.
IMPORTANT: The summary sentence can be incomplete, but ALL the words and phrases in one 
speaker's summary should be from the same speaker in the transcript, not from the context
of other speakers.
\end{alltt}
\end{tcolorbox}
\caption{System Prompt: speaker-aware transcript summarization.}
\label{fig:appendix_summary_prompt}
\end{figure*}

\begin{figure*}[t]

\begin{tcolorbox}[
  colback=white,
  colframe=black,
  boxrule=0.3pt,
  arc=0pt,
  width=\textwidth,
  left=2.5mm,
  right=2.5mm,
  top=1.5mm,
  bottom=1.5mm,
  boxsep=0.8mm
]
\fontsize{8}{9}\selectfont
\begin{alltt}
### TASK DESCRIPTION
You are analyzing a speaker diarization error correction. Your task is to find which part
in the predicted conversation was incorrectly attributed and the correct speaker, based on 
predicted conversation and correction message.
The result contain the following information
- "original_speaker_id" = the original WRONG speaker ID shown in the predicted conversation
bellow.
- "original_sentence" = the exact text with wrong speaker from the conversation bellow.
It can be the a part of the sentence, but it should be a complete phrase that can be 
understood on its own, and covers the information in the correction message.
- "corrected_speaker_id" = the CORRECT speaker ID (who actually spoke those words).
- "corrected_sentence" = the same sentence text, but attributed to the correct speaker.

### INPUT
Given this conversation transcript with incorrect speaker attributions:conversation
And this correction message: text
Available enrolled speakers: enrolled_speakers_str

### OUTPUT FORMAT
Output ONLY this JSON structure:
\{"corrections": [\{
        "original_speaker_id": "wrong speaker from conversation",
        "original_sentence": "exact text from conversation", 
        "corrected_speaker_id": "correct speaker who actually said it",
        "original_sentence": "same text but correctly attributed"\}]\}
        
### EXAMPLE
Example: If conversation shows "A: I like pizza I like baozi" but correction says "A didn't
say she like pizza, B said it", then:
- original_speaker_id: "A" (wrong attribution in conversation)  
- original_sentence: "I like pizza"
- corrected_speaker_id: "B" (who actually said it)
- corrected_sentence: "I like pizza"

### RULES
It is important to copy the exact sentence text from the conversation. Do not paraphrase or
change the text or skip any words. We need to exact match to recognize the correct speaker.  
Ignore sentences that are very short or contain only filler words (e.g., "yeah", "okay",
"right").
Also ignore sentences that clearly contain words from multiple speakers within the same
segment, as these are likely transcription segmentation errors.
IMPORTANT: the original_speaker_id and corrected_speaker_id should both be in the available 
enrolled speakers.     
IMPORTANT: Only output the part of one utterance at the semantic level without punctuation
in correction, can not be the whole utterance. 
DO NOT INCLUDE ANY OTHER TEXT OR COMMENTS.
\end{alltt}
\end{tcolorbox}
\caption{Prompt used for System Correction with user correction feedback into structured speaker correction outputs.}
\label{fig:appendix_prompt}
\end{figure*}

\begin{figure*}[t]

\begin{tcolorbox}[
  colback=white,
  colframe=black,
  boxrule=0.3pt,
  arc=0pt,
  width=\textwidth,
  left=2.5mm,
  right=2.5mm,
  top=1.5mm,
  bottom=1.5mm,
  boxsep=0.8mm
]
\fontsize{8}{9}\selectfont
\begin{alltt}
### TASK DESCRIPTION
You are trying to identify keyword-level speaker attribution errors in a predicted summary.
The ground truth transcript is the source of truth. The predicted summary is the one that
may need correction.

### RULES
Rules:
- Focus on speaker comparison. Only output the speaker attribution errors. Do not include
any other text or comments.
- Prefer speaker attribution evidence based on distinctive statements, self-reference, 
and context key words, rather than topic overlap alone. Do not rely on high-level semantic
similarity alone.
- Use the speaker id's to identify the speaker and do not use the speaker names suggested
by the transcript.
- Find the candidate sentences in the ground truth transcript. If the speakers of candidate 
sentences contains the predicted speaker, do not generate a correction.
- Ignore filler words, short acknowledgements, and procedural speech (e.g., "okay", "yeah",
"next slide", "let's move on") unless they clearly affect speaker attribution.

### INPUT
GROUND TRUTH CONVERSATION: ground_truth_conversation
SUMMARY OF PREDICTED: summary

### OUTPUT REQUIREMENT
Output format:
Wrong Phrase: <exact minimal phrase from summary>
Wrong Speaker: <speaker_id>
Correct Speaker: <speaker_id> (only give one speaker)

### EXAMPLE
For example, if the summary is "The tool is expensive but easy to use", and the wrong phrase
can be "The tool is expensive", or "The tool is easy to use".

If there is no difference, return "None." only.
Important: If words appear in the ground-truth transcript but not in the predicted summary,
do not make correction.
Important: If the importrant words/information is missing in ground truth transcript, do 
not make correction.
\end{alltt}
\end{tcolorbox}

\caption{Simulation Prompt used for difference analysis.}
\label{fig:appendix_difference_prompt}
\end{figure*}

\begin{figure*}[t]
\begin{tcolorbox}[
  colback=white,
  colframe=black,
  boxrule=0.3pt,
  arc=0pt,
  width=\textwidth,
  left=2.5mm,
  right=2.5mm,
  top=1.5mm,
  bottom=1.5mm,
  boxsep=0.8mm
]
\fontsize{8}{9}\selectfont
\begin{alltt}
### TASK DESCRIPTION
You are simulating a user correction for a speaker diarization system in a multi-speaker 
ASR conversation.
Your task is to verify the proposed speaker attribution errors, generate a brief correction
message if the error is valid.
The "DIFFERENCE ANALYSIS" provides candidate speaker errors.

### VERIFICATION RULES
Before generating a correction message, verify:
1. If the important words mentioned in the predicted summary do NOT appear in the ground
truth transcript, do NOT generate a correction.
2. If the proposed phrase is mentioned by the original predicted speaker in ground truth,
do NOT generate a correction. 
3. A correction should only be generated if the predicted content can be linked to one
specific ground-truth speaker more strongly than to other speakers.
If none of the proposed differences are valid speaker attribution errors, return: None.

### OUTPUT REQUIREMENT
Generate some short informal correction messages that:
- Starts with "Hey COBI"
- Mentions the wrong speaker in the predicted summary
- Mentions the key topic or keywords
- Identifies the correct speaker who actually said it
- Is concise and written as a single sentence

Correct all the valid sentences. Do not include any extra commentary or explanation.

### INPUT
GROUND TRUTH CONVERSATION: ground_truth_conversation
SUMMARY OF PREDICTED: summary
DIFFERENCE ANALYSIS: difference_analysis
AVAILABLE SPEAKERS: available_speakers_str
\end{alltt}
\end{tcolorbox}

\caption{Prompt used to simulate user correction messages for verified speaker attribution errors.}
\label{fig:appendix_simulated_correction_prompt}
\end{figure*}

\begin{figure*}[t]
\begin{tcolorbox}[
  colback=white,
  colframe=black,
  boxrule=0.3pt,
  arc=0pt,
  width=\textwidth,
  left=2.5mm,
  right=2.5mm,
  top=1.5mm,
  bottom=1.5mm,
  boxsep=0.8mm
]
\fontsize{8}{9}\selectfont
\begin{alltt}
### TASK DESCRIPTION
Consider the system output below that is for an *automatic ASR correction system*
(based on analyzing a ground-truth transcript and a predicted summary):

### INPUT
GROUND TRUTH TRANSCRIPT 
=======================
ground_truth_conversation

SUMMARY 
=======================
predicted_summary

CORRECTION
=======================
correction_message

## RULES
Now, carefully check the ground truth, especially for the same words in different speakers.
If the key words in correction message appears lots of times in the ground truth transcript
with different speakers, then the correction message is likely incorrect. 
Tell me: is this correction correct or incorrect? 
Only answer "Correct" or "Incorrect". Do not include any other text or explanation.
\end{alltt}
\end{tcolorbox}
\caption{Simulation Verifier Prompt: repeated twice for zero shot.}
\label{fig:appendix_simulated_correction_prompt}
\end{figure*}

\end{document}